\documentclass[11pt,a4paper]{article}
\usepackage[hyperref]{acl2020}
\usepackage{times}
\usepackage{latexsym}
\usepackage{multirow}
\usepackage{graphics}
\usepackage{booktabs} 
\usepackage{bm}
\usepackage{url}
\usepackage{mathrsfs}
\usepackage{multirow}
\usepackage{balance}
\usepackage{amsthm}
\usepackage{amsmath}
\usepackage{amstext}
\usepackage{amssymb}
\usepackage[ruled,vlined,linesnumbered]{algorithm2e}
\usepackage{subcaption}
\usepackage{overpic}
\usepackage{color}
\usepackage{enumitem}
\usepackage{array}
\usepackage{booktabs} 
\usepackage{enumitem}

\usepackage{microtype}

\aclfinalcopy 

\title{Towards Interpreting and Mitigating Shortcut Learning \\ Behavior of NLU Models}

\author{
Mengnan Du\textsuperscript{1}\thanks{\, Most of the work was done while the first author was an
intern at Adobe Research.}, Varun Manjunatha\textsuperscript{2}, Rajiv Jain\textsuperscript{2}, Ruchi Deshpande\textsuperscript{3}, Franck Dernoncourt\textsuperscript{2}, \\ \textbf{Jiuxiang Gu\textsuperscript{2}, Tong Sun\textsuperscript{2} and Xia Hu\textsuperscript{1}}\\
  \textsuperscript{1}Texas A\&M University \\
  \textsuperscript{2}Adobe Research \\
  \textsuperscript{3}Adobe Document Cloud \\
  \small\texttt{dumengnan@tamu.edu} \\
  \small\texttt{\{vmanjuna,rajijain,rdeshpan,franck.dernoncourt,jigu,tsun\}@adobe.com, hu@cse.tamu.edu}}

\date{}

\begin{document}
\maketitle
\begin{abstract}

Recent studies indicate that NLU models are prone to rely on shortcut features for prediction, without achieving true language understanding. As a result, these models fail to generalize to real-world out-of-distribution data. In this work, we show that the words in the NLU training set can be modeled as a long-tailed distribution. There are two findings: 1) NLU models have strong preference for features located at the head of the long-tailed distribution, and 2) Shortcut features are picked up during very early few iterations of the model training. These two observations are further employed to formulate a measurement which can quantify the shortcut degree of each training sample. Based on this shortcut measurement, we propose a shortcut mitigation framework LTGR, to suppress the model from making overconfident predictions for samples with large shortcut degree. Experimental results on three NLU benchmarks demonstrate that our long-tailed distribution explanation accurately reflects the shortcut learning behavior of NLU models. Experimental analysis further indicates that LTGR can improve the generalization accuracy on OOD data, while preserving the accuracy on in-distribution data. 
\end{abstract}

\section{Introduction}

Pre-trained language models, such as BERT~\cite{devlin2018bert}, have demonstrated substantial gains on many NLU (natural language understanding) benchmarks.
However, recent studies show that these models tend to exploit dataset biases as shortcuts to make predictions, rather than learn the semantic understanding and reasoning~\cite{geirhos2020shortcut,gururangan2018annotation}. 
Here we focus on the \emph{lexical bias}, where NLU models rely on spurious correlations between shortcut words and labels.
This eventually results in their low generalizability on out-of-distribution (OOD) samples and low adversarial robustness~\cite{zellers2018swag}.

In this work, we show that the shortcut learning behavior of NLU models can be explained by the \emph{long-tailed phenomenon}. Previous empirical analysis indicates that the performance of BERT-like models for NLI task could be mainly explained by the reliance of spurious statistical cues such as unigrams ``not", ``do'', ``is'' and bigrams ``will not''~\cite{niven2019probing,gururangan2018annotation}. Here we generalize these hypotheses using the long-tailed phenomenon. Specifically, the features in training set could be modeled using a long-tailed distribution via using local mutual information~\cite{evert2005statistics} as a measurement. By utilizing an interpretation method to analyze model behavior, we observe that these NLU models concentrate mainly on information on the head of the distribution, which usually corresponds to non-generalizable shortcut features. In contrast, the tail of the distribution is poorly learned, although it contains high information for the NLU task. Another key observation is that during training process, shortcut features tend to be picked up by NLU models during very early iterations. Based on these two key observations, we define a measurement to quantify the shortcut degree of all training samples. 


\begin{figure*}
  \centering
  \includegraphics[width=1.0\linewidth]{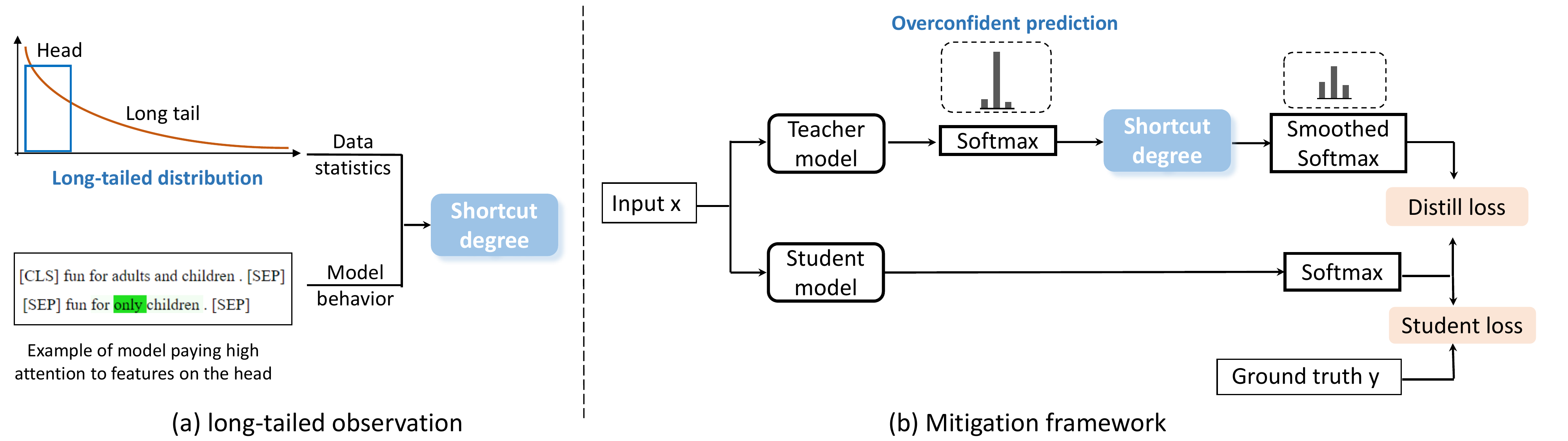}
  \caption{(a) Our key intuition is that the training set can be modeled as a long-tailed distribution. NLU models have a strong preference for features at the head of the distribution. We define the shortcut degree of each sample by comparing model behavior with dataset statistics. (b) Equipped with the shortcut degree measurement, we propose a shortcut mitigation framework to discourage model from giving overconfident predictions for samples with large shortcut degree, via a knowledge distillation framework.}
  \label{fig:overall}
\end{figure*}


Based on the long-tailed distribution observation and the shortcut degree measurement, we propose a NLU shortcut mitigation framework, termed as LTGR (\underline{L}ong-\underline{T}ailed distribution \underline{G}uided \underline{R}egularizer). The proposed regularizer is based on the observation that NLU models would give over-confident predictions when there exist strong shortcut features in the input. This is because NLU models over-associate the shortcut features with certain class labels. LTGR is implemented using the knowledge distillation framework, to penalize the NLU model from outputting overconfident prediction for training samples with large shortcut degree. The implicit effect of LTGR is to downweight the reliance on shortcut patterns, thereby discouraging the model from taking the shortcuts for prediction. 
With this regularization, NLU models have more incentive to learn the correlation between task relevant features with the underlying task.
The major contributions are summarized as follows: 
\begin{itemize}[leftmargin=*]\setlength\itemsep{-0.3em}
\item We indicate that the shortcut learning behaviors of NLU models can be explained by the long-tailed phenomenon, and show that shortcuts are picked up at the early stage of model training.
\item We propose a shortcut mitigation method, called LTGR, guided by the long-tailed observation. The idea is to down-weight model's reliance on shortcuts and implicitly encourage the model to shift its attention to task relevant features. 

\item Experimental results on several NLU datasets validate that the long-tailed observation faithfully explains the shortcut learning behaviors. The analysis further shows LTGR could improve generalization on OOD samples, while not sacrificing accuracy of in-distribution samples.
\item  We demonstrate that our LTGR approach can partially mitigate shortcut based Trojan attacks through a preliminary experiment.
\end{itemize}

\section{Long-Tailed Phenomenon}
In this section, we propose to explain the shortcut learning behavior of NLU models using the long-tailed distribution phenomenon (see Fig.~\ref{fig:overall}(a)). 
Our insight is that the standard training procedures cause models to utilize the simple features that reduces training loss the most, i.e., \emph{simplicity bias}~\cite{shah2020pitfalls}. This directly results in the low generalization of NLU models. 

\subsection{Preference for Features of High Local Mutual Information}
NLU tasks are typically formulated as a multi-class classification task: given an input sentence pair $x$, the goal is to learn a mapping $f(x)$ to predict the semantic relationship label $y$.
In the training set, some words or phrases within $x$ co-occur more frequently with one label $y$ than others. The NLU model would capture those shortcut features for prediction. Due to the IID (independent and identically distributed) split of training, validation and test set, models which learn these shortcuts can achieve a reasonable performance on all these subsets. Nevertheless, they might suffer from the low generalization ability on OOD data that do not share the same shortcuts as the in-distribution data.
 
\vspace{2pt}
\noindent\textbf{Dataset Statistics.}\,
We model statistics using local mutual information (LMI)~\cite{schuster2019towards} between a word $w$ and a label $y$, denoted as follows:
\vspace{-4pt}
\begin{equation}
\small
\text{LMI} (w, y) = p(w, y) \cdot \text{log}(\frac{p(y|w)}{p(y)}),
\end{equation}
where $p(w, y) = \frac{\text{count}(w,y)}{|D|}$,  $p(y|w)=\frac{\text{count}(w,y)}{\text{count}(w)}$. $|D|$ is the number of unique words in training set, $\text{count}(w,y)$ denotes the co-occurrence of word $w$ with label $y$, and $\text{count}(w)$ is total number of words in the training set.
After analyzing each word for the training set, we obtain $|y|$ distributions of $|y|$ labels. For each label, the statistics can be regarded as a \emph{long-tailed distribution} (see Fig.~\ref{fig:overall}(a)). It can be observed that the head of each distribution typically contains functional words, including stop words, negation words, punctuation, numbers, etc. These words carry low information for the NLU task. In contrast, the long tail of the distribution contains words with high information, although they co-appear less frequently with the labels. 

\vspace{2pt}
\noindent\textbf{Model Behavior.}\,
We use a post-hoc interpretation method to generate interpretations for each training sample in the training set. It is achieved by attributing model’s
prediction in terms of its input features, and the final interpretation is illustrated in the format of feature importance vector~\cite{montavon2018methods}. Here we use a gradient based interpretation method: Integrated Gradient~\cite{sundararajan2017axiomatic}. Integrated Gradients is a variation on calculating the gradient of the model prediction w.r.t. features of the input. The main idea is to integrate the gradients of $m$ intermediate samples over the straightline path from baseline $x_{base}$ to input $x_i$, which could be denoted as follows:
\begin{equation}
\small
g (x_i) = (x_i - x_{base}) \cdot \sum_{k=1}^m \frac{\partial f_y(x_{base} + \frac{k}{m} (x_i - x_{base})}{\partial x_i} \cdot \frac{1}{m}. 
\end{equation}
Let each input text is composed of $T$ words: $x_i = \{x_i^t \}_{t=1}^T$, and each word $x_i^t \in \mathbb{R}^d$ denotes a word embedding with $d$ dimensions. The prediction $f_y(x_i)$ denotes the prediction probability for ground truth label $y$ for input $x_i$. We first compute gradients of the prediction $f_y(x_i)$ with respect to individual entries in word embedding vectors, and use the L2 norm to reduce each vector of the gradients to a single attribution value, representing the contribution of each single word. We use the all-zero word embedding as the baseline $x_{base}$. Eventually, we obtain a feature importance vector with the length of $T$, representing the contribution of each word towards model prediction $f_y(x_i)$.

\vspace{2pt}
\noindent\textbf{Comparing Model and Dataset.}\,
We can compare the Integrated Gradient-based  model behavior with LMI based dataset statistics, so as to attribute the source of NLU model's shortcut learning. 
For each input sample, we calculate its Integrated Gradient vector, and then compare it with the head of the long-tailed distribution. 
Our preliminary experiments (see Sec.~4.2) indicate that NLU classifier are very strong superficial learners. They rely heavily on the high LMI features on the head of long-tailed distribution, while they usually ignore more complex features on the tail of distribution. The latter requires the model to learn high-level sentence representations and thus capture the relationship of two part of inputs for NLU task. Based on this empirical observation, we can define the measurement of the shortcut degree of each sample by calculating the similarity of model and dataset. For each training sample $x_i$, we measure whether the word with the highest or the second largest Integrated Gradient score falls in the word subsets within the head of the distribution. Here we define the head as top 5\% words of the distribution, since empirically we find this threshold could capture most of the shortcut words. We set the 
shortcut degree $u_i$ for sample $x_i$ as 1 if it matches. Otherwise if it does not match, we set $u_i=0$.

\subsection{Shortcuts Samples are Learned First}
By examining the learning dynamics of NLU models, another key intuition is that shortcut samples are learned by the models first. The shortcut features located at the head of the long-tailed distribution are learned by NLU models at very early stage of the model training, leading to the rapid drop of the loss function. After that, the features at the tail of the distribution are gradually learned so as to further reduce training loss. Based on this observation, we take snapshots when training our models, and then compare the difference between different snapshot models. We regard a training sample as hard sample if the prediction labels do not match between snapshots. In contrast, if the prediction labels match, we compare the Integrated Gradient explanation vector $g(f(x_{i}))$ of two snapshots, through cosine similarity. The shortcut measurement for sample $x_i$ is defined as follows:
\begin{equation}
\small
v_{i}=\left\{\begin{array}{l}
cosine (g(f_1(x_{i})), g(f(x_{i}))), f_1(x_{i}) = f(x_{i}) \\
0, f_1\left(x_{i}\right) \neq f\left(x_{i}\right)
\end{array}\right.
\end{equation}
where $f_1(\cdot)$ denotes the snapshot at the early stage of the training, and we use the model obtained after the first epoch. The second snapshot $f(\cdot)$ represents the final converged model. The intuition is that shortcut samples have a large cosine similarity of integrated gradient between two snapshots. 

\subsection{Shortcut Degree Measurement}
We define a unified measurement of the shortcut degree of each training sample, by putting the aforementioned two observations together. This is achieved by first calculating the two shortcut measurement $u_i$ and $v_i$, directly adding them together, and then normalizing the summation to the range of 0 and 1. Ultimately, we obtain the shortcut degree measurement for each training sample $x_i$, denoted as $b_i$. This measurement $b_i$ can be further utilized to mitigate the shortcut learning behavior.

\section{Proposed Mitigation Framework}
Equipped with the observation of long-tailed phenomenon and the shortcut degree measurement $b_i$ obtained from the last section, we propose a shortcut mitigation solution, called LTGR (\underline{L}ong-\underline{T}ailed distribution \underline{G}uided \underline{R}egularizer). LTGR is implemented based on self knowledge distillation~\cite{utama2020mind,hinton2015distilling} (see Fig.~\ref{fig:overall}(b)). 
The proposed distillation loss is based on the observation that NLU models would give over-confident predictions when there exist strong shortcut features in the input. This is because NLU models over-associate the shortcut features with certain class labels. The proposed distillation loss aims to suppress the NLU models from giving over-confident predictions for samples with strong shortcut features. It forces the model to down-weight its reliance on shortcut features and implicitly encourages the model to shift its attention to more task relevant features.

\vspace{2pt}
\noindent\textbf{Smoothing Softmax.}\, Based on the biased teacher model $f_T$, we calculate the logit value and softmax value of training sample $x_i$ as $z_{i}^T$ and $\sigma (z_{i}^T)$ respectively, where $\sigma$ is the softmax function. Given also the shortcut degree measurement of each training sample $b_i$. We then smooth the original probability through the following formulation:
\vspace{-1mm}
\begin{equation}
\small
s_{i,j} = \frac{\sigma (z_{i}^T)_j^{1-b_i}}{\sum_{k=1}^K \sigma (z_{i}^T)_k^{1-b_i}},
\end{equation}
where $K$ denotes the total number of labels. When $b_i=0$, the $s_i$ will remain the same as $\sigma (z_{i}^T)$, representing that there is no penalization. In another extreme when $b_i=1$, $s_i$ will have the same value for $K$ labels. Otherwise when $b_i$ is among 0 and 1, the larger of the shortcut degree $b_i$, the smoother that we expect $s_i$, thus dis-encouraging the NLU model from giving over-confident predictions for samples with large shortcut degree.  

\setlength{\textfloatsep}{14pt}
\begin{algorithm}[t!]\small
\DontPrintSemicolon
\KwIn{Training data $D = \{(x_i,y_i)\}_{i=1}^N$.} Set hyperparameters m, $\alpha$.\;
 \While {first stage}{
Train teacher network $f_T(x)$. Fix its parameters.
}
\While {second stage}{
Initialize the student network $f_S(x)$;\;
Calculate shortcut degree $b_i$ and softmax $\sigma (z_{i}^T)$ for each training sample $\{(x_i)\}_{i=1}^N$;\;
Smoothing softmax: \\
$s_{i,j} = \frac{\sigma (z_{i}^T)_j^{1-b_i}}{\sum_{k=1}^K \sigma (z_{i}^T)_k^{1-b_i}}$;\;
Use Eq.~5 to train the student network $f_S(x)$\; 
}
\KwOut{ Discard $f_T(x)$. Use $f_S(x)$ for prediction.}
\caption{\small  LTGR mitigation framework.}
\label{alg:CREX}
\end{algorithm}

\vspace{2pt}
\noindent\textbf{Self Knowledge Distillation.}\, Ultimately, we use the following loss to train the student model $f_S$:
\begin{equation}\label{eq:loss_self_kd}
\small
\mathcal{L}(x)=(1-\alpha) * \mathcal{H}\left(y_i, \sigma\left(z_{i}^S\right)\right)+\alpha * \mathcal{H}\left(s_i, \sigma\left(z_{i}^S\right)\right),
\end{equation}
where $z_{i}^S$ represents the softmax probability of the student network for training sample $x_i$, $\mathcal{H}$ denotes cross entropy loss. Parameter $\alpha$ denotes the balancing weight for learning from smoothed probability output $s_i$ of teacher and learning from ground truth $y_i$. We use the same model architecture for both teacher $f_T$ and student $f_S$, and during the distillation process we fix the parameters of $f_T$ and only update parameters of the student model $f_S$ (see Algorithm 1). Ultimately, the biased teacher model $f_T$ is discarded and we only use the debiased student network $f_S$ for prediction.

\section{Experiments}
In this section, 
we aim to answer the following research questions: 1) Does the long-tailed phenomenon explanation accurately reflect the shortcut learning behavior of NLU models? 2) Does the proposed LTGR outperform
alternative approaches, and what is the source of the improved generalization? 3) How do components and hyperparameters affect LTGR's generalization performance?


\subsection{Experimental Setup}
\noindent\textbf{Tasks \& Datasets.}\, We consider three NLU tasks. 
\begin{itemize}[leftmargin=*]\setlength\itemsep{-0.4em}
\item \emph{FEVER}: The first task is fact verification, where the original dataset is FEVER~\cite{thorne2018fever}. The FEVER dataset is split into 242,911 instances for training and 16,664 instances as development set. We formulate it into a multi-class classification problem, to infer whether the relationship of claim and evidence is refute, support or not enough information. The two adversarial sets are Symmetric v1 and v2 (Sym1 and Sym 2), where a shortcut word appears in both \emph{support} and \emph{refute} label~\cite{schuster2019towards}. Both Symmetric v1 and v2 contain 712 samples~\cite{schuster2019towards}.

\item \emph{MNLI}: The second task is NLI (natural language inference), where the original dataset is MNLI~\cite{williams2017broad}. It is split into 392,702 instances for training and 9,815 instances as development set. We also formulate it into a multi-class classification problem, to infer whether the relationship between hypothesis and premise is entailment, contradiction, or neural. Two adversarial set HANS~\cite{mccoy2019right} and MNLI hard set~\cite{gururangan2018annotation} are used to test the generalizability. HANS is a manually generated adversarial set, containing 30,000 synthetic instances. Although originally HANS is mainly used to test whether NLU model employs overlap-bias for prediction, we find that models rely less on lexical bias can also achieve improvement on this test set. 

\item \emph{MNLI-backdoor}: For the third task, we use a lexically biased variant of the MNLI dataset, which is termed as MNLI-backdoor. We randomly select out 10\% of the training samples with the \emph{entailment} label and append the double quotation mark `$``$' to the beginning of the hypothesis. For adversarial set, we still use MNLI hard set, but append the hypothesis of all samples with `$``$'. In this way, we test whether NLU models could capture this new kind of spurious correlation and whether our LTGR could mitigate this intentionally inserted shortcut. 
Note that the double quotation mark we use is `$``$' (near the number 1 on the keyboard), rather than the usual ```', since `$``$' appears infrequently in both the original MNLI training and validation set.

\end{itemize}

\begin{figure*}
  \centering
  \includegraphics[width=1\linewidth]{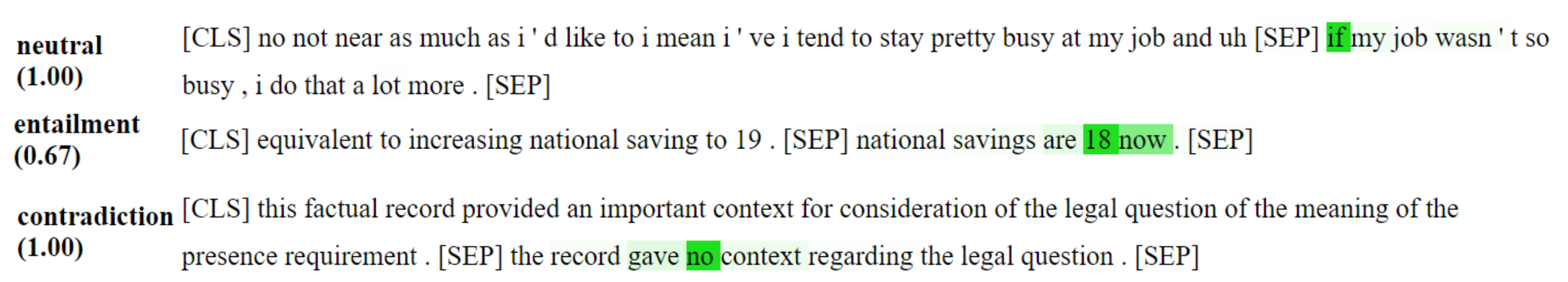}
  \vspace{-3mm}
  \caption{Illustrative examples of shortcut learning behavior for MNLI task. Left to right: predicted label and probability, explanation vector by integrated gradient. Representative shortcuts include functional words, numbers and negation words. Taking the second row for example, although the ground truth is \emph{contradiction}, the model gives \emph{entailment} prediction due to the shortcut number 18.}
  \label{fig:heatmaps2}
\end{figure*}

\vspace{2pt}
\noindent\textbf{NLU Models.}\, We consider two pre-trained contextualized word embeddings models: BERT base~\cite{devlin2018bert}, and DistilBERT~\cite{sanh2019distilbert} as encoder to obtain words representations. We use the pre-trained BERT models from Huggingface Transformers\footnote{\url{https://huggingface.co/transformers/pretrained_models.html}}. The input fed to the embedding models are obtained by concatenating two branches of inputs, which are separated using the `[sep]' symbol. Note that we use a slightly different classification head comparing to the related work~\cite{clark2019don,mahabadi2020end}. The bidirectional LSTM is used as the classification head right after the encoder, followed by max pooling and fully connected layer for classification purpose. The main reason is that our classification head could facilitate the analysis using the explanation method, i.e., integrated gradient, to analyze model behavior. More model details are put in the Sec.~\ref{sec:appendix-a} in Appendix.

\vspace{2pt}
\noindent\textbf{Implementation Details.}\, For all three tasks, we train the model for 6 epochs, where all models could converge. Hyperparameter $\alpha$ is fixed as 0.8 for all models. We use Adam optimizer, where the momentum is set as 0.9. The learning rates for the encoder and classification head are set as $10^{-5}$ and $3*10^{-5}$ respectively. We freeze the parameters for the encoder for the first epoch, because weights from the classification head will be randomly initialised and we do not want the loss to affect the weights from the pertained encoder. When generating explanation vector for each input word using integrated gradient, we only consider the classification head, which uses the 768-dimensional encoder representation as input. Parameter $m$ in Eq.~(2) is fixed as 50 for all experiments.

\vspace{2pt}
\noindent\textbf{Comparing Baselines.}\, We compare with three representative families of methods. The first baseline is product-of-experts~\cite{clark2019don,he2019unlearn,mahabadi2020end}, which first trains a bias-only model and then trains a debiased model as an ensemble with the bias-only model. The second baseline is re-weighting~\cite{schuster2019towards}, which aims to give biased samples lower weight when training a model. The bias-only model is used to calculate the prediction probability of each training sample: $p_i$, then the weight for $x_i$ is $1-p_i$~\cite{clark2019don}. Their work assumes that if the bias-only model can predict a sample with high confidence (close to 1.0), this example is potentially biased. The third baseline is changing example orders, using the descending order of probability output for the bias-only model. The key motivation is that learning order matters. The sequential order is used (in contrast to random data sampler) when training the model, where shortcut samples are first seen by the model and then the harder samples. Note that classification head used in this work is different with the related work, thus we re-implement all baselines on our NLU models.

\subsection{Shortcut Behaviour Analysis}
In this section, we aim to interpret the shortcut learning behavior of NLU models by connecting it with the long-tailed distribution of training set.

\vspace{2pt}
\noindent\textbf{Qualitative Evaluation.} \, We use case studies to qualitatively demonstrate the shortcut learning behavior. Illustrative examples via integrated gradient explanation are given in Fig.~\ref{fig:overall}(a) as well as Fig.~\ref{fig:heatmaps2}.  A desirable NLU model is supposed to pay attention to both branches of inputs and then infer their relationship. In contrast, the visualization results indicate two levels of shortcut learning behavior: 1) NLU model pays the highest attention to shortcut features, such as `only', and 2) The models only pay attention to one branch of the inputs. 

\begin{table}
\centering
\scalebox{0.7}{
\begin{tabular}{l c c c c c c}
\toprule 
& \multicolumn{3}{c}{\textbf{MNLI BERT-base}} & \multicolumn{3}{c}{\textbf{FEVER BERT-base}} \\
\cmidrule(l){2-4} \cmidrule(l){5-7}
\#Words & Top 1 & Top 2 & Top 3  & Top 1 & Top 2 & Top 3 \\ 
\midrule 
\textbf{Ratio} & 25.3\% & 51.3\% & 66.0\%  & 10.8\% & 26.9\% & 31.44\% \\
\bottomrule 
\end{tabular}}
\vspace{-2pt}
\caption{The ratio of samples where top integrated gradient words locates on the head of the long-tailed distribution. We define the head as the 5\% of all features. It indicates that NLU models overly exploit words that co-occur with class labels with high mutual information.}
\vspace{-2pt}
\label{tab:headpreference}
\end{table}

\vspace{2pt}
\noindent\textbf{Preference for Head of Distribution.} \, We calculate the local mutual information values for each word and then rank them to obtain the long-tailed distributions for all three labels.  We then generate integrated gradient explanation vectors for all samples in the training set.  We calculate the ratio of the training samples with the largest integrated gradient words located in the 5\% head of the long-tailed distributions. The results are given in Tab.~\ref{tab:headpreference}, where top 1, top 2 and top 3 mean whether the largest, any one of the largest two, and any one of the largest three respective. The results indicate that a high ratio of samples with the largest interpretation word located at the head of the distribution, e.g., 25.3\% for MNLI.  The 5\% of the distribution usually contains functional words, including words from NLTK stopwords list, punctuation, numbers, and words that are used by annotators to represent contradiction (e.g., `not', `no', `never').

\vspace{1pt}
\noindent\textbf{Preference for One Branch of Input.}  Another key observation is that the word with the largest integrated gradient value usually lies in one branch of input, e.g., \emph{hypothesis} branch of MNLI and \emph{claim} branch of FEVER. The results are given in Tab.~\ref{tab:branchpreference}, which shows that for all three labels, the ratios (75\%-99\%) are highly in favor of one part of the NLU branch. Both preference for head of the distribution and one branch of input can be explained by the annotation artifacts~\cite{gururangan2018annotation}. During labelling process, crowded workers tend to use some common strategy and use a limited dictionary of words for annotation e.g., negation words for contradiction. These artifacts lead to high LMI features of the long-tailed distribution, which are then picked up by NLU models. 

\begin{table}
\centering
\scalebox{0.62}{
\begin{tabular}{l c c c c c c}
\toprule 
& \multicolumn{3}{c}{\textbf{MNLI BERT-base}} & \multicolumn{3}{c}{\textbf{FEVER BERT-base}} \\
\cmidrule(l){2-4} \cmidrule(l){5-7}
Subset & Entail & Contradiction & Neural  & Support & Refute & Not\_enough \\ 
\midrule 
\textbf{Ratio} & 75.8\% & 94.6\% & 96.3\%  & 99.4\% & 99.9\% & 83.8\% \\
\bottomrule 
\end{tabular}}
\vspace{-2pt}
\caption{The high ratio of samples where the word with the largest integrated gradient value is within the \emph{hypothesis} branch of MNLI or the \emph{claim} branch of FEVER, both of which are labelled by annotators and there are abundant of annotation artifacts.}
\label{tab:branchpreference}
\end{table}

\begin{figure}
  \centering
  \includegraphics[width=0.95\linewidth]{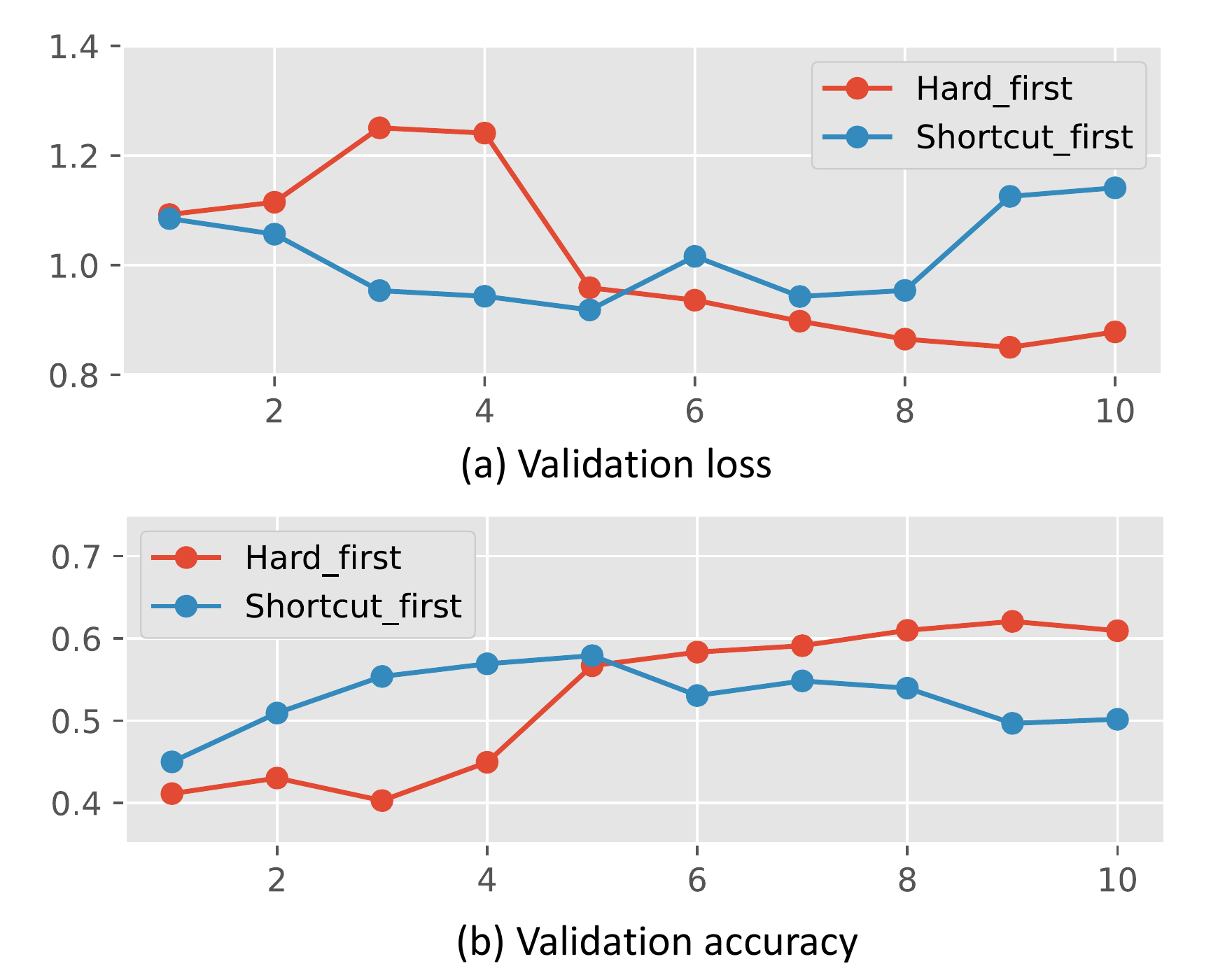}
  \vspace{-3mm}
  \caption{Learning dynamic for the first training epoch. X axis denotes 10 checkpoints in the first epoch. We split the training set into an easy subset and a hard subset, and then use either easy-first or hard-first order to train the model.  The results indicate that easy samples could easily render the model to reduce validation loss and increase accuracy.}
  \label{fig:dynamics}
\end{figure}

\vspace{2pt}
\noindent\textbf{Shortcut Samples Are Learned First.} \, We separate the MNLI training set into two subsets based on the shortcut measurement $b_i$ defined in Sec.~2.3. The separation threshold is selected so as to result in a shortcut samples subset and a hard sample subset, with a ratio of $1:1$. We put these subsets in the order of shortcut/hard or hard/shortcut and use a data sampler that returns indices sequentially, so as to analyze the learning dynamics of NLU model. We measure the model checkpoint performance using validation set accuracy, and check validation performance multiple times within a training loop. Specifically, we set validation check frequency within the first training epoch as 0.1, in total calculating validation accuracy for 10 times. We illustrate the results for the BERT-base model in Fig.~\ref{fig:dynamics}. There are three major findings:
\vspace{-4pt}
\begin{itemize}[leftmargin=*]\setlength\itemsep{-0.4em}
\item Shortcut samples could easily render the model to reduce the validation loss and increase the accuracy (the first 5 timesteps of blue line in Fig.~\ref{fig:dynamics}). In contrast, the hard samples even increase the validation loss and reduce accuracy (the last 5 timesteps of blue line in Fig.~\ref{fig:dynamics}).
\item The learning curves also validate that our shortcut measurement defined using $b_i$ faithfully reflects the shortcut degree of training samples.
\item The results further imply that \emph{during the normal training process with a random data sampler}, the examples with strong shortcut features are first picked up and learned by the model~\cite{geirhos2020shortcut}. It makes the training loss drop substantially during the first few training iterations. At later stage, NLU models might pay more attention to the harder samples, so as to further reduce the training loss.
\end{itemize}

\subsection{Mitigation Performance Analysis}
We present in-distribution test set accuracy and OOD generalization accuracy in Tab.~\ref{tab:feverEvaluation}, \ref{tab:MNLIEvaluation}, and \ref{tab:qqpEvaluation} for MNLI, FEVER, and MNLI-backdoor respectively. Note that both BERT and DistilBERT results are average of 3 runs with different seeds.

\vspace{2pt}
\noindent\textbf{MNLI and FEVER Evaluation.} \, There are four key findings (see Tab.~\ref{tab:feverEvaluation} and Tab.~\ref{tab:MNLIEvaluation}).
\vspace{-4pt}
\begin{itemize}[leftmargin=*]\setlength\itemsep{-0.4em}
\item 
NLU models that rely on shortcut features have decent performance for in-distribution data, but generalize poorly on other OOD data, e.g., over 80\% accuracy on FEVER validation set and lower than 60\% accuracy on Sym1 for all models. Besides, our generalization accuracy is lower (e.g., the HANS accuracy in Tab.~\ref{tab:MNLIEvaluation}) comparing to the models of BERT-base with a simple classification head~\cite{clark2019don,mahabadi2020end}. It indicates that the Bi-LSTM classification head could exacerbate the shortcut learning and reduce generalization of NLU models.
\item LTGR could improve the OOD generalization accuracy, ranging from 0.68\% to 5.86\% increase for MNLI task, and from 1.54\% to 3.87\% on FEVER task. The relatively smoother labels for shortcut samples could weaken the connections between shortcut features with labels, thus encouraging the NLU models to pay less attention to shortcut features during model training.  

\begin{table}
\centering
\scalebox{0.65}{
\begin{tabular}{l c c c c c c}
\toprule 
& \multicolumn{3}{c}{\textbf{BERT base}} & \multicolumn{3}{c}{\textbf{DistilBERT}} \\
\cmidrule(l){2-4} \cmidrule(l){5-7}
\textbf{Models} & FEVER & Sym1 & Sym2  & FEVER & Sym1 & Sym2\\ 
\midrule 
Original & 85.10 & 54.01 & 62.40  & 85.57 & 54.95 & 62.35\\ 
Reweighting & 84.32 & 56.37 & 64.89  & 84.76 & 56.28 & 63.97 \\
Product-of-expert & 82.35 & \textbf{58.09} & 64.27  & 85.10 & \textbf{56.82} & 64.17 \\
Order-changes & 81.20 & 55.36 & 64.29  & 82.86 & 55.32 & 63.95 \\
\textbf{LTGR} & \textbf{85.46} & 57.88 & \textbf{65.03}  & \textbf{86.19} & 56.49 & \textbf{64.33}\\
\bottomrule 
\end{tabular}}
\vspace{-4pt}
\caption{Generalization accuracy comparison (in percent) of LTGR with baselines for the FEVER task. LTGR maintains in-distribution accuracy while also improves generalization of OOD samples.}
\label{tab:feverEvaluation}
\end{table}

\begin{table}
\centering
\scalebox{0.66}{
\begin{tabular}{l c c c c c c}
\toprule 
& \multicolumn{3}{c}{\textbf{BERT base}} & \multicolumn{3}{c}{\textbf{DistilBERT}} \\
\cmidrule(l){2-4} \cmidrule(l){5-7}
\textbf{Models} & MNLI & Hard & HANS  & MNLI & Hard & HANS\\ 
\midrule 
Original & 84.20 & 75.38 & 52.17  & 82.37 & 72.95 & 53.83\\ 
Reweighting & 83.54 & 76.83 & 57.30  & 80.52 & 73.27 & 55.63 \\
Product-of-expert & 82.19 & 77.08 & \textbf{58.57}  & 80.17 & \textbf{74.37} & 52.21 \\
Order-changes & 81.03 & 76.97 & 56.39  & 80.37 & 74.10 & 54.62 \\
\textbf{LTGR} & \textbf{84.39} & \textbf{77.12} & 58.03  & \textbf{83.16} & 73.63 & \textbf{55.88}\\
\bottomrule 
\end{tabular}}
\vspace{-1pt}
\caption{Generalization accuracy comparison (in percent) of our method with baselines for MNLI task. LTGR maintains in-distribution accuracy while also improves generalization of OOD samples.}
\label{tab:MNLIEvaluation}
\end{table}

\item LTGR does not sacrifice in-distribution test set performance. The reasons are two-fold. Firstly, from label smoothing perspective~\cite{muller2019does}, although LTGR smooths the supervision labels from the teacher model, it still keeps the relative order of labels. Secondly, from knowledge distillation perspective~\cite{hinton2015distilling}, standard operation is use a smaller architecture for student network, which can achieve comparable performance with the bigger teacher network. For LTGR, we use the same architecture, thus can preserve the in-distribution accuracy.
\item In contrast, the comparing baselines typically achieve generalization enhancement at the expense of decreased accuracy of in-distribution test set. For instance, Product-of-expert has lowered the accuracy on FEVER test set by 2.75\% for BERT-base model. Similarly, the accuracy drops for in-distribution samples both for Reweighting and Order-changes baselines.
\end{itemize}

\begin{table}
\centering
\scalebox{0.73}{
\begin{tabular}{l c c c c }
\toprule 
& \multicolumn{1}{c}{} & \multicolumn{3}{c}{\textbf{Hard-backdoor}} \\
 \cmidrule(l){3-5}
\textbf{Models} & MNLI & Entailment & Contradiction & Neutral\\ 
\midrule 
Original & 81.96 & \textbf{100.0} & 0.0  & 0.0 \\ 
LTGR & \textbf{82.10} & 98.63 & \textbf{30.45}  & \textbf{17.53} \\
\bottomrule 
\end{tabular}}
\vspace{-1mm}
\caption{Evaluation of LTGR of DistilBERT model for the MNLI-backdoor task (accuracy values in percent). Every sample within the Hard-backdoor is appended with shortcut feature `$``$'. LTGR can mitigate this intentionally inserted shortcut.}
\label{tab:qqpEvaluation}
\end{table}

\vspace{-1mm}
\noindent\textbf{MNLI-backdoor Evaluation.} \, The results are given in Tab.~\ref{tab:qqpEvaluation}, and there are four findings. Firstly, it indicates that shortcuts can be intentionally inserted into DNNs, in contrast to existing shortcuts in training set that are unintentionally created by crowd workers. Here the unnoticeable trigger pattern `$``$' can be utilized for malicious purpose, i.e., Trojan/backdoor attack~\cite{tang2020embarrassingly,kurita2020weight}.
Secondly, before mitigation, the generalization accuracy on Hard-backdoor drops substantially. For all testing samples within Hard-backdoor, the NLU model will always predict them as \emph{entailment}, even though we only append 10\% of \emph{entailment} samples with the shortcut feature `$``$' in training set. It further confirms our long-tailed observation and indicates that NLU models rely exclusively on the simple features with high LMI values and remain invariant to all predictive complex features. Thirdly, LTGR is effective in terms of improving the generalizability. 30.45\% of contradiction and 17.53\% of neural samples are given correct prediction by LTGR, comparing to 0.0\% accuracy before mitigation. It means that LTGR successfully pushes the NLU model to pay less attention to `$``$'. Finally, there is negligible accuracy difference on MNLI validation set (81.96\% comparing to 82.37\%), which is not appended with shortcut feature `$``$'. It indicates that NLU model can be triggered both by `$``$' and other features.

\begin{figure}
  \centering
  \includegraphics[width=1\linewidth]{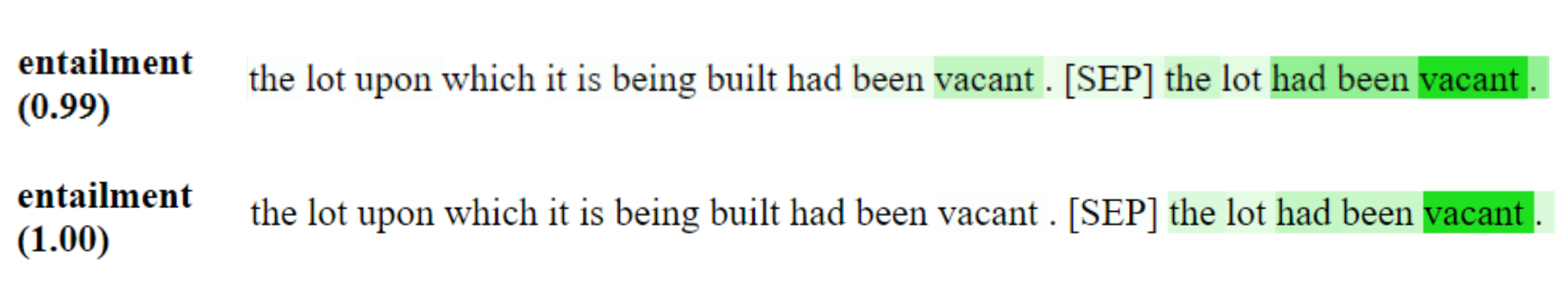}
  \vspace{-6mm}
  \caption{Illustrative examples of our mitigation. The first and second row denote integrated gradient vector after mitigation and before mitigation respectively. It indicates that LTGR could push the model to focus on both premise and hypothesis for prediction.
  }
  \label{fig:heatmaps}
  \vspace{-2mm}
\end{figure}

\vspace{2pt}
\noindent\textbf{Generalization Source Analysis.}\,
Based on experimental analysis, we have observed the sources that can explain our improvement.
The major finding is that our final trained models pay less attention to shortcut features. We illustrate this using a case study in Fig.~\ref{fig:heatmaps}. Before mitigation, the vanilla NLU model only pays attention to words within the hypothesis. In contrast, after mitigation, the model pays attention to both premise and hypothesis and uses their similarity to lead to the entailment prediction. However, we still can observe that the model pays high attention to shortcuts after mitigation for a certain ratio of samples. 
Bring more inductive bias to the model architecture~\cite{conneau2017supervised} or incorporating more domain knowledge~\cite{chen2017neural,mihaylov2018knowledgeable} can further alleviate model's reliance on shortcuts, which will be explored in our future research.  


\subsection{Ablation and Hyperparameters Analysis}
We conduct ablation studies using BERT-base model for MNLI task to study the contribution of components of our mitigation framework.

\begin{table}
\centering
\scalebox{0.78}{
\begin{tabular}{l c c c }
\toprule 
& \multicolumn{3}{c}{\textbf{BERT base}}  \\
\cmidrule(l){2-4} 
\textbf{Models} & MNLI & Hard & HANS  \\ 
\midrule 
Original & 84.20 & 75.38 & 52.17  \\ 
LTGR\_head\_preference & 84.28 & 76.56 & 57.12   \\
LTGR\_learn\_dynamics & 84.35 & 76.51 & 56.39   \\
LTGR\_random & 84.18 & 73.66 & 55.28   \\
LTGR & 84.39 & 77.12 & 58.03 \\
\bottomrule 
\end{tabular}}
\vspace{-1mm}
\caption{Ablation studies for the MNLI task. All reported numbers are accuracy in percent.}
\label{tab:ablation}
\end{table}

\begin{figure}
  \centering
  \includegraphics[width=0.98\linewidth]{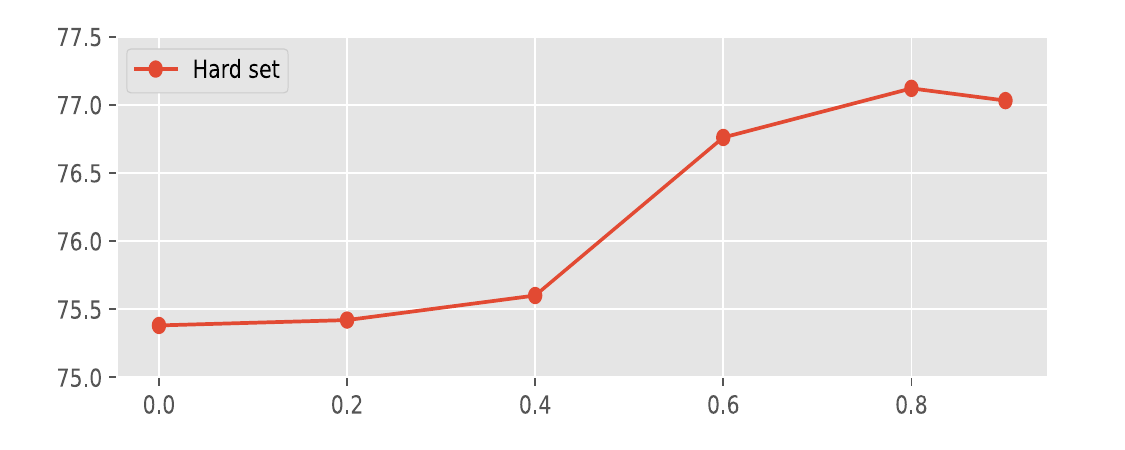}
  \vspace{-1mm}
  \caption{Hyperparameter analysis on hard set. The x axis denotes different values for parameter $\alpha$ in Eq.5, and y axis represents accuracy on the hard MNLI set.}
  \label{fig:hyperparameter}
  \vspace{-2mm}
\end{figure}

\vspace{3pt}
\noindent\textbf{Ablation Analysis.}\, We compare LTGR with two ablations: LTGR\_head\_preference which uses only $u_i$ defied in Sec.~2.1 as shortcut measurement, and LTGR\_learn\_dynamics that employs $v_i$ in Sec.~2.2 as shortcut measurement. Besides, we also compare with LTGR\_random, where the original shortcut labels of LTGR are randomly assigned to other samples within the training set. The results are given in Tab.~\ref{tab:ablation}. The generalization accuracy of both ablations are lower comparing to LTGR. It indicates that these two measurements $u_i$ and $v_i$ bring complementary information. Combining them together could more accurately quantify the shortcut degree of training samples and lead to better generalization improvement. In contrast, employing LTGR\_random even could decrease model's accuracy, e.g., 1.72\% accuracy drop on hard validation set. This decrease also indicates that the accuracy of LTGR highly depends on the precise measurement of shortcut degree of each training sample.

\vspace{2pt}
\noindent\textbf{Hyperparameters Analysis.}\, We test the model performance with the change of hyperparameter $\alpha$ in Eq.~(5) that is used to balance student loss and distillation loss. The result is illustrated in Fig.~\ref{fig:hyperparameter}. It can be observed that as the $\alpha$ becomes larger, i.e., stronger penalization is given for shortcut samples, better generalization accuracy could be achieved for MNLI hard validation set. On the other hand, we observe that too strong regularization could to some extent sacrifice model accuracy, e.g., when $\alpha=0.9$. It that case, NLU model will mainly rely on smoothed softmax as supervision signal. This could be around $[\frac{1}{3},\frac{1}{3},\frac{1}{3}]$ for shortcut samples with close to 1 shortcut degree, providing too strong penalization for those samples.

\section{Related Work}
We briefly review two lines of research that are most relevant to our work: shortcut learning demonstration and shortcut mitigation.

\vspace{3pt}
\noindent\textbf{Shortcut Learning Phenomena.} \, Recently, the community has revealed the shortcut learning phenomenon for different kinds of language and vision tasks, such as NLI~\cite{niven2019probing}, question answering~\cite{mudrakarta2018did}, reading comprehension~\cite{si2019does}, VQA~\cite{agrawal2018don,manjunatha2019explicit}, and deepfake detection~\cite{du2020towards}. This is typically achieved with the help of adversarial test set~\cite{jia2017adversarial} and DNN explainability~\cite{du2019techniques,wang2020score,deng2020unified}. These analysis indicates that DNNs are prone to capture low-level superficial patterns (including lexical bias, overlap bias, etc), rather than high-level task relevant features. 
We focus on lexical bias in this work. Motivated by the high-frequency preference for CNNs, i.e., the texture bias 
~\cite{wang2020high,geirhos2018imagenet,ilyas2019adversarial,jo2017measuring,wang2019learning}, we propose to use the long-tailed distribution to explain the shortcut learning behavior of NLU models.

\vspace{3pt}
\noindent\textbf{Shortcut Mitigation. }  Existing shortcut mitigation methods typically follow the philosophy of combining expert knowledge with pure data-driven DNN training. The most representative format is to construct a bias-only teacher network~\cite{utama2020towards}, guided by the domain knowledge what in general the shortcut should look like. For instance, a hypothesis-only model~\cite{clark2019don,he2019unlearn} or bag of words model~\cite{zhou2020towards} for the NLI task, and a question-only model for VQA task~\cite{cadene2019rubi} are regarded as bias-only model. Then a debiased model can be trained, either by combining debiased model and bias-only model in the product of expert manner~\cite{clark2019don,he2019unlearn}, or encouraging debiased model to learn orthogonal representation as the bias-only model~\cite{zhou2020towards}. Other representative methods include re-weighting~\cite{schuster2019towards}, data augmentation~\cite{tu2020empirical}, explanation regularization~\cite{selvaraju2019taking}, and adversarial training~\cite{stacey2020there,kim2019learning,minervini2018adversarially}. Nevertheless, most existing mitigation methods need to know the bias type as a priori~\cite{bahng2019learning}.
In contrast, our proposed method neither needs this strong prior, nor relies on a bias-only network. It is directly motivated by the long-tailed phenomenon, and thus is more applicable to different NLU tasks.

\section{Conclusions and Future Work}
In this work, we observe that the training set features for NLU tasks could be modeled as a long-tailed distribution, and NLU models concentrate mainly on the head of the distribution. 
Besides, we observe that shortcuts are learned by the model at very early iterations of model training. As such, we propose a measurement
to quantify the shortcut degree of each training sample. Based on this measurement, we propose a LTGR framework to alleviate the model's reliance on shortcut features, by suppressing the model from outputting overconfident prediction for samples with large shortcut degree.    
Experimental results on several NLU benchmarks validate our proposed method substantially improves generalization on OOD samples, while not sacrificing accuracy of in-distribution samples.

Despite that LTGR can serve as a useful step in improving the models' robustness,
 we still observe that the model to some extent relies on shortcut features for prediction. Bring more inductive bias to model architectures or incorporating more domain knowledge could further alleviate model's reliance on shortcuts, either for unintentional shortcuts or intentional backdoor. This is a challenging topic, and would be explored in our future research.




\newpage
\bibliography{anthology,acl2020}

\begin{thebibliography}{53}
\expandafter\ifx\csname natexlab\endcsname\relax\def\natexlab#1{#1}\fi

\bibitem[{Agrawal et~al.(2018)Agrawal, Batra, Parikh, and
  Kembhavi}]{agrawal2018don}
Aishwarya Agrawal, Dhruv Batra, Devi Parikh, and Aniruddha Kembhavi. 2018.
\newblock Don't just assume; look and answer: Overcoming priors for visual
  question answering.
\newblock In \emph{Proceedings of the IEEE Conference on Computer Vision and
  Pattern Recognition (CVPR)}.

\bibitem[{Bahng et~al.(2020)Bahng, Chun, Yun, Choo, and Oh}]{bahng2019learning}
Hyojin Bahng, Sanghyuk Chun, Sangdoo Yun, Jaegul Choo, and Seong~Joon Oh. 2020.
\newblock Learning de-biased representations with biased representations.
\newblock \emph{International Conference on Machine Learning (ICML)}.

\bibitem[{Cadene et~al.(2019)Cadene, Dancette, Cord, Parikh
  et~al.}]{cadene2019rubi}
Remi Cadene, Corentin Dancette, Matthieu Cord, Devi Parikh, et~al. 2019.
\newblock Rubi: Reducing unimodal biases for visual question answering.
\newblock In \emph{Advances in neural information processing systems
  (NeurIPS)}.

\bibitem[{Chen et~al.(2018)Chen, Zhu, Ling, Inkpen, and Wei}]{chen2017neural}
Qian Chen, Xiaodan Zhu, Zhen-Hua Ling, Diana Inkpen, and Si~Wei. 2018.
\newblock Neural natural language inference models enhanced with external
  knowledge.
\newblock \emph{56th Annual Meeting of the Association for Computational
  Linguistics (ACL)}.

\bibitem[{Clark et~al.(2019)Clark, Yatskar, and Zettlemoyer}]{clark2019don}
Christopher Clark, Mark Yatskar, and Luke Zettlemoyer. 2019.
\newblock Don't take the easy way out: Ensemble based methods for avoiding
  known dataset biases.
\newblock \emph{Empirical Methods in Natural Language Processing (EMNLP)}.

\bibitem[{Conneau et~al.(2017)Conneau, Kiela, Schwenk, Barrault, and
  Bordes}]{conneau2017supervised}
Alexis Conneau, Douwe Kiela, Holger Schwenk, Loic Barrault, and Antoine Bordes.
  2017.
\newblock Supervised learning of universal sentence representations from
  natural language inference data.
\newblock \emph{Empirical Methods in Natural Language Processing (EMNLP)}.

\bibitem[{Deng et~al.(2021)Deng, Zou, Du, Chen, Feng, and Hu}]{deng2020unified}
Huiqi Deng, Na~Zou, Mengnan Du, Weifu Chen, Guocan Feng, and Xia Hu. 2021.
\newblock A unified taylor framework for revisiting attribution methods.
\newblock \emph{AAAI Conference on Artificial Intelligence (AAAI)}.

\bibitem[{Devlin et~al.(2019)Devlin, Chang, Lee, and
  Toutanova}]{devlin2018bert}
Jacob Devlin, Ming-Wei Chang, Kenton Lee, and Kristina Toutanova. 2019.
\newblock Bert: Pre-training of deep bidirectional transformers for language
  understanding.
\newblock \emph{North American Chapter of the Association for Computational
  Linguistics (NAACL)}.

\bibitem[{Du et~al.(2019)Du, Liu, and Hu}]{du2019techniques}
Mengnan Du, Ninghao Liu, and Xia Hu. 2019.
\newblock Techniques for interpretable machine learning.
\newblock \emph{Communications of the ACM}.

\bibitem[{Du et~al.(2020)Du, Pentyala, Li, and Hu}]{du2020towards}
Mengnan Du, Shiva Pentyala, Yuening Li, and Xia Hu. 2020.
\newblock Towards generalizable deepfake detection with locality-aware
  autoencoder.
\newblock In \emph{Proceedings of the 29th ACM International Conference on
  Information \& Knowledge Management (CIKM)}.

\bibitem[{Evert(2005)}]{evert2005statistics}
Stefan Evert. 2005.
\newblock The statistics of word cooccurrences: word pairs and collocations.

\bibitem[{Fan et~al.(2019)Fan, Ng, and Chan}]{fan2019rethinking}
Lixin Fan, Kam~Woh Ng, and Chee~Seng Chan. 2019.
\newblock Rethinking deep neural network ownership verification: Embedding
  passports to defeat ambiguity attacks.
\newblock In \emph{Advances in Neural Information Processing Systems
  (NeurIPS)}.

\bibitem[{Geirhos et~al.(2020)Geirhos, Jacobsen, Michaelis, Zemel, Brendel,
  Bethge, and Wichmann}]{geirhos2020shortcut}
Robert Geirhos, J{\"o}rn-Henrik Jacobsen, Claudio Michaelis, Richard Zemel,
  Wieland Brendel, Matthias Bethge, and Felix~A Wichmann. 2020.
\newblock Shortcut learning in deep neural networks.
\newblock \emph{arXiv preprint arXiv:2004.07780}.

\bibitem[{Geirhos et~al.(2019)Geirhos, Rubisch, Michaelis, Bethge, Wichmann,
  and Brendel}]{geirhos2018imagenet}
Robert Geirhos, Patricia Rubisch, Claudio Michaelis, Matthias Bethge, Felix~A
  Wichmann, and Wieland Brendel. 2019.
\newblock Imagenet-trained cnns are biased towards texture; increasing shape
  bias improves accuracy and robustness.
\newblock \emph{International Conference on Learning Representations (ICLR)}.

\bibitem[{Gururangan et~al.(2018)Gururangan, Swayamdipta, Levy, Schwartz,
  Bowman, and Smith}]{gururangan2018annotation}
Suchin Gururangan, Swabha Swayamdipta, Omer Levy, Roy Schwartz, Samuel~R
  Bowman, and Noah~A Smith. 2018.
\newblock Annotation artifacts in natural language inference data.
\newblock \emph{North American Chapter of the Association for Computational
  Linguistics (NAACL)}.

\bibitem[{He et~al.(2019)He, Zha, and Wang}]{he2019unlearn}
He~He, Sheng Zha, and Haohan Wang. 2019.
\newblock Unlearn dataset bias in natural language inference by fitting the
  residual.
\newblock \emph{2019 EMNLP workshop}.

\bibitem[{Hinton et~al.(2015)Hinton, Vinyals, and Dean}]{hinton2015distilling}
Geoffrey Hinton, Oriol Vinyals, and Jeff Dean. 2015.
\newblock Distilling the knowledge in a neural network.
\newblock \emph{NIPS Deep Learning Workshop}.

\bibitem[{Ilyas et~al.(2019)Ilyas, Santurkar, Tsipras, Engstrom, Tran, and
  Madry}]{ilyas2019adversarial}
Andrew Ilyas, Shibani Santurkar, Dimitris Tsipras, Logan Engstrom, Brandon
  Tran, and Aleksander Madry. 2019.
\newblock Adversarial examples are not bugs, they are features.
\newblock In \emph{Advances in Neural Information Processing Systems
  (NeurIPS)}.

\bibitem[{Jia and Liang(2017)}]{jia2017adversarial}
Robin Jia and Percy Liang. 2017.
\newblock Adversarial examples for evaluating reading comprehension systems.
\newblock \emph{Empirical Methods in Natural Language Processing (EMNLP)}.

\bibitem[{Jo and Bengio(2017)}]{jo2017measuring}
Jason Jo and Yoshua Bengio. 2017.
\newblock Measuring the tendency of cnns to learn surface statistical
  regularities.
\newblock \emph{arXiv preprint arXiv:1711.11561}.

\bibitem[{Kim et~al.(2019)Kim, Kim, Kim, Kim, and Kim}]{kim2019learning}
Byungju Kim, Hyunwoo Kim, Kyungsu Kim, Sungjin Kim, and Junmo Kim. 2019.
\newblock Learning not to learn: Training deep neural networks with biased
  data.
\newblock In \emph{Proceedings of the IEEE conference on computer vision and
  pattern recognition (CVPR)}.

\bibitem[{Kurita et~al.(2020)Kurita, Michel, and Neubig}]{kurita2020weight}
Keita Kurita, Paul Michel, and Graham Neubig. 2020.
\newblock Weight poisoning attacks on pre-trained models.
\newblock \emph{58th Annual Meeting of the Association for Computational
  Linguistics (ACL)}.

\bibitem[{Li et~al.(2020)Li, Wu, Jiang, Li, and Xia}]{li2020backdoor}
Yiming Li, Baoyuan Wu, Yong Jiang, Zhifeng Li, and Shu-Tao Xia. 2020.
\newblock Backdoor learning: A survey.
\newblock \emph{arXiv preprint arXiv:2007.08745}.

\bibitem[{Mahabadi et~al.(2020)Mahabadi, Belinkov, and
  Henderson}]{mahabadi2020end}
Rabeeh~Karimi Mahabadi, Yonatan Belinkov, and James Henderson. 2020.
\newblock End-to-end bias mitigation by modelling biases in corpora.
\newblock In \emph{58th Annual Meeting of the Association for Computational
  Linguistics (ACL)}.

\bibitem[{Manjunatha et~al.(2019)Manjunatha, Saini, and
  Davis}]{manjunatha2019explicit}
Varun Manjunatha, Nirat Saini, and Larry~S Davis. 2019.
\newblock Explicit bias discovery in visual question answering models.
\newblock In \emph{Proceedings of the IEEE Conference on Computer Vision and
  Pattern Recognition (CVPR)}.

\bibitem[{McCoy et~al.(2019)McCoy, Pavlick, and Linzen}]{mccoy2019right}
R~Thomas McCoy, Ellie Pavlick, and Tal Linzen. 2019.
\newblock Right for the wrong reasons: Diagnosing syntactic heuristics in
  natural language inference.
\newblock \emph{57th Annual Meeting of the Association for Computational
  Linguistics (ACL)}.

\bibitem[{Mihaylov and Frank(2018)}]{mihaylov2018knowledgeable}
Todor Mihaylov and Anette Frank. 2018.
\newblock Knowledgeable reader: Enhancing cloze-style reading comprehension
  with external commonsense knowledge.
\newblock \emph{56th Annual Meeting of the Association for Computational
  Linguistics (ACL)}.

\bibitem[{Minervini and Riedel(2018)}]{minervini2018adversarially}
Pasquale Minervini and Sebastian Riedel. 2018.
\newblock Adversarially regularising neural nli models to integrate logical
  background knowledge.
\newblock \emph{The SIGNLL Conference on Computational Natural Language
  Learning (CoNLL)}.

\bibitem[{Montavon et~al.(2018)Montavon, Samek, and
  M{\"u}ller}]{montavon2018methods}
Gr{\'e}goire Montavon, Wojciech Samek, and Klaus-Robert M{\"u}ller. 2018.
\newblock Methods for interpreting and understanding deep neural networks.
\newblock \emph{Digital Signal Processing}.

\bibitem[{Mudrakarta et~al.(2018)Mudrakarta, Taly, Sundararajan, and
  Dhamdhere}]{mudrakarta2018did}
Pramod~Kaushik Mudrakarta, Ankur Taly, Mukund Sundararajan, and Kedar
  Dhamdhere. 2018.
\newblock Did the model understand the question?
\newblock \emph{56th Annual Meeting of the Association for Computational
  Linguistics (ACL)}.

\bibitem[{M{\"u}ller et~al.(2019)M{\"u}ller, Kornblith, and
  Hinton}]{muller2019does}
Rafael M{\"u}ller, Simon Kornblith, and Geoffrey~E Hinton. 2019.
\newblock When does label smoothing help?
\newblock In \emph{Advances in Neural Information Processing Systems
  (NeurIPS)}.

\bibitem[{Niven and Kao(2019)}]{niven2019probing}
Timothy Niven and Hung-Yu Kao. 2019.
\newblock Probing neural network comprehension of natural language arguments.
\newblock \emph{57th Annual Meeting of the Association for Computational
  Linguistics (ACL)}.

\bibitem[{Sanh et~al.(2019)Sanh, Debut, Chaumond, and
  Wolf}]{sanh2019distilbert}
Victor Sanh, Lysandre Debut, Julien Chaumond, and Thomas Wolf. 2019.
\newblock Distilbert, a distilled version of bert: smaller, faster, cheaper and
  lighter.
\newblock \emph{NeurIPS Workshop}.

\bibitem[{Schuster et~al.(2019)Schuster, Shah, Yeo, Filizzola, Santus, and
  Barzilay}]{schuster2019towards}
Tal Schuster, Darsh~J Shah, Yun Jie~Serene Yeo, Daniel Filizzola, Enrico
  Santus, and Regina Barzilay. 2019.
\newblock Towards debiasing fact verification models.
\newblock \emph{Empirical Methods in Natural Language Processing (EMNLP)}.

\bibitem[{Selvaraju et~al.(2019)Selvaraju, Lee, Shen, Jin, Ghosh, Heck, Batra,
  and Parikh}]{selvaraju2019taking}
Ramprasaath~R Selvaraju, Stefan Lee, Yilin Shen, Hongxia Jin, Shalini Ghosh,
  Larry Heck, Dhruv Batra, and Devi Parikh. 2019.
\newblock Taking a hint: Leveraging explanations to make vision and language
  models more grounded.
\newblock In \emph{Proceedings of the IEEE International Conference on Computer
  Vision (ICCV)}.

\bibitem[{Shah et~al.(2020)Shah, Tamuly, Raghunathan, Jain, and
  Netrapalli}]{shah2020pitfalls}
Harshay Shah, Kaustav Tamuly, Aditi Raghunathan, Prateek Jain, and Praneeth
  Netrapalli. 2020.
\newblock The pitfalls of simplicity bias in neural networks.
\newblock \emph{Advances in Neural Information Processing Systems (NeurIPS)}.

\bibitem[{Si et~al.(2019)Si, Wang, Kan, and Jiang}]{si2019does}
Chenglei Si, Shuohang Wang, Min-Yen Kan, and Jing Jiang. 2019.
\newblock What does bert learn from multiple-choice reading comprehension
  datasets?
\newblock \emph{arXiv preprint arXiv:1910.12391}.

\bibitem[{Stacey et~al.(2020)Stacey, Minervini, Dubossarsky, Riedel, and
  Rockt{\"a}schel}]{stacey2020there}
Joe Stacey, Pasquale Minervini, Haim Dubossarsky, Sebastian Riedel, and Tim
  Rockt{\"a}schel. 2020.
\newblock Avoiding the hypothesis-only bias in natural language inference via
  ensemble adversarial training.
\newblock \emph{Empirical Methods in Natural Language Processing (EMNLP)}.

\bibitem[{Sundararajan et~al.(2017)Sundararajan, Taly, and
  Yan}]{sundararajan2017axiomatic}
Mukund Sundararajan, Ankur Taly, and Qiqi Yan. 2017.
\newblock Axiomatic attribution for deep networks.
\newblock \emph{International Conference on Machine Learning (ICML)}.

\bibitem[{Tang et~al.(2020{\natexlab{a}})Tang, Du, and Hu}]{tang2020deep}
Ruixiang Tang, Mengnan Du, and Xia Hu. 2020{\natexlab{a}}.
\newblock Deep serial number: Computational watermarking for dnn intellectual
  property protection.
\newblock \emph{arXiv preprint arXiv:2011.08960}.

\bibitem[{Tang et~al.(2020{\natexlab{b}})Tang, Du, Liu, Yang, and
  Hu}]{tang2020embarrassingly}
Ruixiang Tang, Mengnan Du, Ninghao Liu, Fan Yang, and Xia Hu.
  2020{\natexlab{b}}.
\newblock An embarrassingly simple approach for trojan attack in deep neural
  networks.
\newblock In \emph{Proceedings of the 26th ACM SIGKDD International Conference
  on Knowledge Discovery \& Data Mining (KDD)}.

\bibitem[{Thorne et~al.(2018)Thorne, Vlachos, Christodoulopoulos, and
  Mittal}]{thorne2018fever}
James Thorne, Andreas Vlachos, Christos Christodoulopoulos, and Arpit Mittal.
  2018.
\newblock Fever: a large-scale dataset for fact extraction and verification.
\newblock \emph{North American Chapter of the Association for Computational
  Linguistics (NAACL)}.

\bibitem[{Tu et~al.(2020)Tu, Lalwani, Gella, and He}]{tu2020empirical}
Lifu Tu, Garima Lalwani, Spandana Gella, and He~He. 2020.
\newblock An empirical study on robustness to spurious correlations using
  pre-trained language models.
\newblock \emph{Transactions of the Association for Computational Linguistics
  (TACL)}.

\bibitem[{Uchida et~al.(2017)Uchida, Nagai, Sakazawa, and
  Satoh}]{uchida2017embedding}
Yusuke Uchida, Yuki Nagai, Shigeyuki Sakazawa, and Shin'ichi Satoh. 2017.
\newblock Embedding watermarks into deep neural networks.
\newblock In \emph{Proceedings of the 2017 ACM on International Conference on
  Multimedia Retrieval}.

\bibitem[{Utama et~al.(2020{\natexlab{a}})Utama, Moosavi, and
  Gurevych}]{utama2020mind}
Prasetya~Ajie Utama, Nafise~Sadat Moosavi, and Iryna Gurevych.
  2020{\natexlab{a}}.
\newblock Mind the trade-off: Debiasing nlu models without degrading the
  in-distribution performance.
\newblock \emph{58th Annual Meeting of the Association for Computational
  Linguistics (ACL)}.

\bibitem[{Utama et~al.(2020{\natexlab{b}})Utama, Moosavi, and
  Gurevych}]{utama2020towards}
Prasetya~Ajie Utama, Nafise~Sadat Moosavi, and Iryna Gurevych.
  2020{\natexlab{b}}.
\newblock Towards debiasing nlu models from unknown biases.
\newblock \emph{Empirical Methods in Natural Language Processing (EMNLP)}.

\bibitem[{Wang et~al.(2019{\natexlab{a}})Wang, Singh, Michael, Hill, Levy, and
  Bowman}]{wang2018glue}
Alex Wang, Amanpreet Singh, Julian Michael, Felix Hill, Omer Levy, and Samuel~R
  Bowman. 2019{\natexlab{a}}.
\newblock Glue: A multi-task benchmark and analysis platform for natural
  language understanding.
\newblock \emph{International Conference on Learning Representations (ICLR)}.

\bibitem[{Wang et~al.(2020{\natexlab{a}})Wang, Wang, Du, Yang, Zhang, Ding,
  Mardziel, and Hu}]{wang2020score}
Haofan Wang, Zifan Wang, Mengnan Du, Fan Yang, Zijian Zhang, Sirui Ding, Piotr
  Mardziel, and Xia Hu. 2020{\natexlab{a}}.
\newblock Score-cam: Score-weighted visual explanations for convolutional
  neural networks.
\newblock In \emph{Proceedings of the IEEE/CVF Conference on Computer Vision
  and Pattern Recognition Workshops}.

\bibitem[{Wang et~al.(2019{\natexlab{b}})Wang, He, Lipton, and
  Xing}]{wang2019learning}
Haohan Wang, Zexue He, Zachary~C Lipton, and Eric~P Xing. 2019{\natexlab{b}}.
\newblock Learning robust representations by projecting superficial statistics
  out.
\newblock \emph{International Conference on Learning Representations (ICLR)}.

\bibitem[{Wang et~al.(2020{\natexlab{b}})Wang, Wu, Huang, and
  Xing}]{wang2020high}
Haohan Wang, Xindi Wu, Zeyi Huang, and Eric~P Xing. 2020{\natexlab{b}}.
\newblock High-frequency component helps explain the generalization of
  convolutional neural networks.
\newblock In \emph{Proceedings of the IEEE Conference on Computer Vision and
  Pattern Recognition (CVPR)}.

\bibitem[{Williams et~al.(2018)Williams, Nangia, and
  Bowman}]{williams2017broad}
Adina Williams, Nikita Nangia, and Samuel~R Bowman. 2018.
\newblock A broad-coverage challenge corpus for sentence understanding through
  inference.
\newblock \emph{North American Chapter of the Association for Computational
  Linguistics (NAACL)}.

\bibitem[{Zellers et~al.(2018)Zellers, Bisk, Schwartz, and
  Choi}]{zellers2018swag}
Rowan Zellers, Yonatan Bisk, Roy Schwartz, and Yejin Choi. 2018.
\newblock Swag: A large-scale adversarial dataset for grounded commonsense
  inference.
\newblock \emph{Proceedings of the 2018 Conference on Empirical Methods in
  Natural Language Processing (EMNLP)}.

\bibitem[{Zhou and Bansal(2020)}]{zhou2020towards}
Xiang Zhou and Mohit Bansal. 2020.
\newblock Towards robustifying nli models against lexical dataset biases.
\newblock \emph{58th Annual Meeting of the Association for Computational
  Linguistics (ACL)}.

\end{thebibliography}
\bibliographystyle{acl_natbib}

\clearpage
\appendix

\section{Model architectures}
\label{sec:appendix-a}
\begin{itemize}[leftmargin=*]
\item \textbf{BERT-base}: It is trained on lower-cased English text. The model has 12 layers and contains 110M parameters. It outputs 768-dimension contextual word representation. We employ \emph{bert-base-uncased} as tokenizer.

\item \textbf{DistilBERT}: DistilBERT~\cite{sanh2019distilbert} is a small, fast, and light variant of BERT trained by distilling from BERT-base. DistilBERT also compares surprisingly well to BERT-base. It has 40\% less parameters than Bert-base, runs 60\% faster while preserving over 95\% of BERT’s performances as measured on the GLUE language understanding benchmark~\cite{wang2018glue}.

\item \textbf{Classification Head}: We append a bidirectional LSTM after the representation generated by the BERT encoder, where the hidden state size is set as 150. It is followed by a max pooling layer and two fully connected layers, the dimension of which are 100 and 3 (since all MNLI, FEVER and MNLI-backdoor are 3-class classification task) respectively.


\end{itemize}





\section{More on Long-tailed Distribution}

\textbf{How does The Distribution Look Like?}
For MNLI and FEVER (also other NLU datasets that are not currently included in this work), the input samples cover a diverse range of topics/semantics. Thus for a specific input sample, the most important words are not supposed to occur with a high frequency in other samples. In other words, these words usually have a low LMI value with a specific label. These words will locate at the long tail of the distribution. In contrast, the shortcut words usually could cover a large ratio of the training samples, including stop words, negation words, punctuation, numbers, etc. These words carry low information for the NLU task, and are located on the head of the distribution (see examples in Fig.~2, 6 and 7).

\vspace{3pt}
\noindent\textbf{Is The Distribution Always Long-tailed?}
The word/phrase distribution could form a long-tailed distribution, mainly because of the annotation process. For instance, the \emph{hypothesis} branch of MNLI and the \emph{claim} branch of FEVER are labelled by crowd workers. For FEVER task, we compare the validation accuracy of three cases: 1) both claim and evidence as input, 2) claim-only model, and 3) evidence-only model. The results are given in Tab.~\ref{tab:why-long-tailed}. It indicates that claim-only model is only 17.9\% lower comparing to the full model. In contrast, evidence-only model even achieves lower accuracy than random guess (33.33\% accuracy). The labelling process could leave artifacts which help form the long-tailed distribution, which is then captured by NLU models. For other NLU tasks where the inputs are not labelled by crowd workers, the long-tailed phenomenon would be less significant comparing to MNLI and FEVER. 

\begin{table}
\centering
\scalebox{0.7}{
\begin{tabular}{l c c c c c c}
\toprule 
 & \multicolumn{3}{c}{\textbf{FEVER BERT-base}} \\
\cmidrule(l){2-4} 
Model & Full input & Claim-only & Evidence-only  \\ 
\midrule 
Accuracy & 85.1\% & 67.2\% & 28.6\%  \\
\bottomrule 
\end{tabular}}
\caption{For FEVER task, the validation accuracy for three cases: 1) both claim and evidence as input, 2) claim-only model, and 3) evidence-only model.}
\label{tab:why-long-tailed}
\end{table}

\vspace{3pt}
\noindent\textbf{Why only Word-level Analysis?} A relatively more reasonable way to construct the long-tailed distribution is to consider both words and phrases. Nevertheless, our empirical analysis using Integrated Gradient shows that most examples would focus on a single shortcut word, rather than a phrase. Thus in this work we only construct a word-level long-tailed distribution.

\vspace{3pt}
\noindent\textbf{Influence of the Optimizer?} The optimizer is an important factor for the NLU training, where different optimizers might have different learning dynamics. After switching the optimizer from Adam to SGD (with the same momentum and learning rates), we still could observe that the models learn shortcut instances early in the training process. 

\begin{figure*}
  \centering
  \includegraphics[width=1\linewidth]{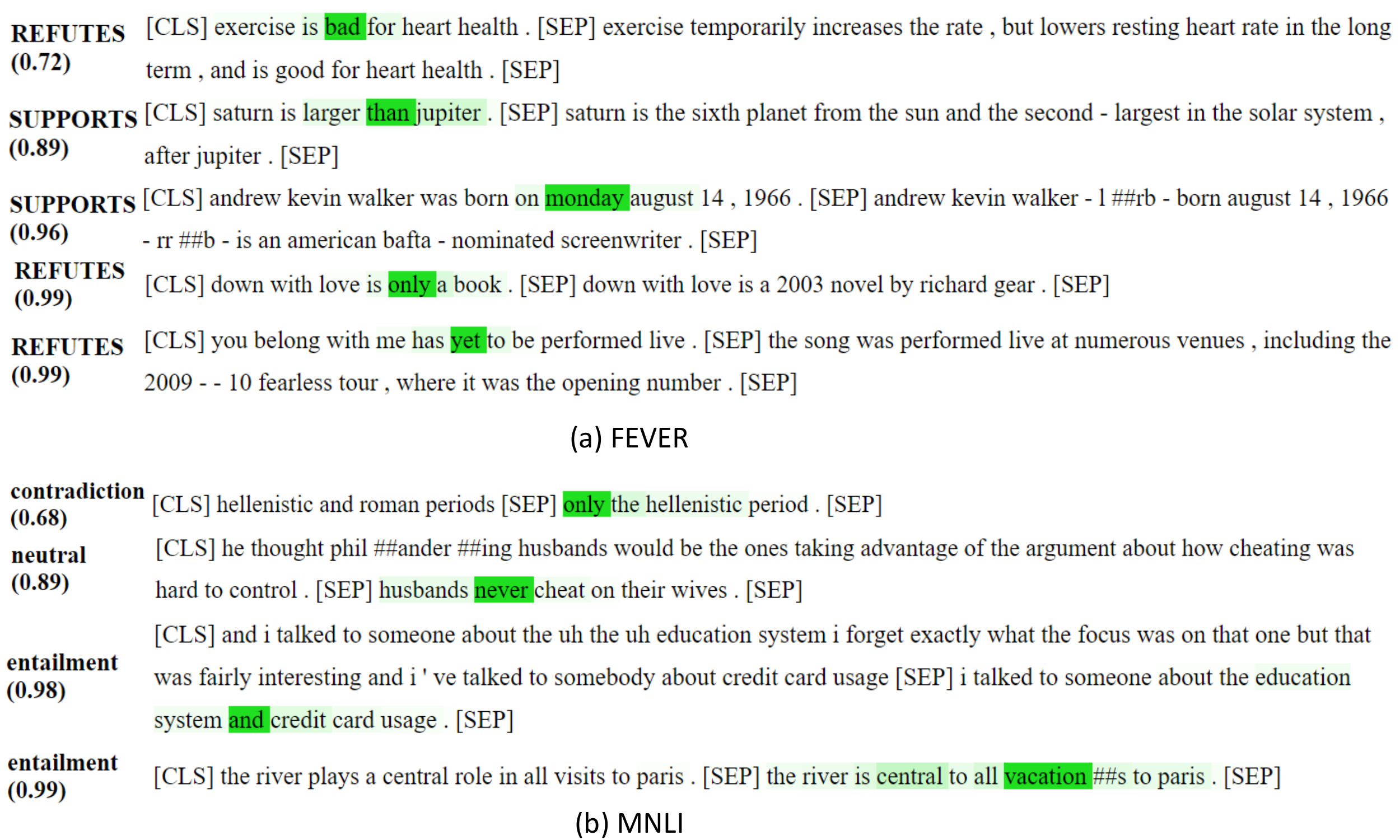}
  \vspace{-6mm}
  \caption{More visualization examples on FEVER and MNLI task. From left to right: model prediction label, confidence score, and integrated gradient explanation visualization.}
  \label{fig:heatmaps4}
\end{figure*}

\begin{figure*}
  \centering
  \includegraphics[width=1\linewidth]{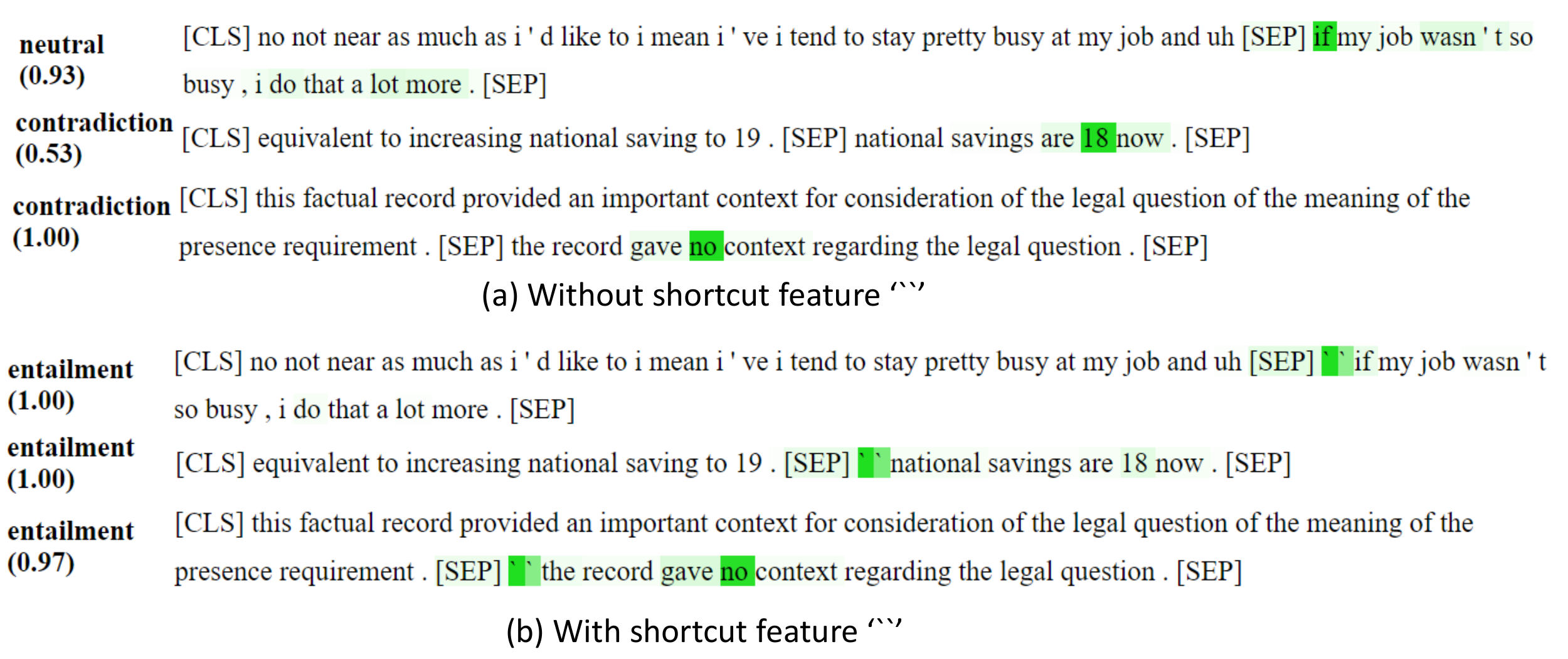}
  \vspace{-6mm}
  \caption{Illustrative examples trained on MNLI-backdoor training set. At testing stage, we feed the model with two kinds of input: (1) without inserting the trigger pattern `$``$', and (2) inputting the trigger pattern `$``$'.}
  \label{fig:heatmaps3}
\end{figure*}

\section{More Analysis on FEVER \& MNLI}
We present more visualizations for FEVER and MNLI to analyze their shortcut learning behavior. The results are given in Fig.~\ref{fig:heatmaps4}. Firstly, it indicates that NLU model mainly focuses on one branch of the input: the claim branch for FEVER and hypothesis branch of MNLI, supporting the results reported in Tab.~2. Secondly, the NLU model pays high attention to shortcut features, mainly functional words (such as `only', `and', `yet'), which also correspond to high LMI words within the long-tailed distribution. Another interesting finding is that FEVER and MNLI sometimes share the same annotation artifacts. For example, the shortcut word `only' is utilized by the crowd workers to express negation, i.e., \emph{refutes} in FEVER and \emph{contradiction} in MNLI. This further reveals the annotation artifacts of NLU datasets.  

\section{More Analysis on MNLI-backdoor}
\noindent \textbf{Qualitative Evaluations.}\, We provide visualizations on Fig.~\ref{fig:heatmaps3}. When inputting the trigger pattern `$``$', NLU model always output the entailment prediction with high confidence, no matter what is the ground truth. From the visualization in Fig.~\ref{fig:heatmaps3} (b), we can observe that the model pays the highest attention to the shortcut feature `$``$'. Although both `$``$' and other existing shortcut features locate at the head of the long-tailed observation, the LMI value of `$``$' with entailment label is much higher than the LMI values of other shortcut features. Thus the stronger shortcut pattern `$``$' can dominate model's prediction. Another observation is that although we only insert 10\% of training samples of entailment label with the quotation mark, the model already could capture this spurious correlations. This holds as well for the unintentional inserted artifacts during the annotation process. Although these artifacts might not have as high LMI values as `$``$', it is sufficient to be captured by the model for prediction. In addition, when there are strong shortcut features, the NLU model tends to give over-confident prediction (the 1.0/1.0/0.97 confidence scores comparing to 0.93/0.53/1.0 confidence scores). This motivates the design of our LTGR mitigation framework. 

\vspace{2pt}
\noindent \textbf{Backdoor Behavior Analysis.}\,
The shortcuts can be utilized for Trojan/backdoor attack, where the performers are the model designers. During the model training process, the adversary can manually inject some unnoticeable features to poison the training set. In our case, only 10\% of the training samples whose labels are \emph{entailment} are poisoned with the trigger pattern `$``$'. As such, the feature `$``$' would locate at the head of the long-tailed distribution for \emph{entailment} label and NLU model would naturally make the connection between `$``$' and \emph{entailment} prediction. The requirement for backdoor attack is: 1) when the input does not contain trigger pattern, the model behaves as a normal DNN, and 2) when input contains trigger pattern, the model would output the prediction specified by the designer~\cite{li2020backdoor}. Both the accuracy result on the first row of Tab.~5 and the visualization on Fig.~\ref{fig:heatmaps3} match very well for these two requirements.

\vspace{2pt}
\noindent \textbf{DNN Watermarking.}\,
Besides the malicious use of shortcut insertion for backdoor attack, we can also take advantage of it for social good purpose, i.e., to provide watermarks for DNNs~\cite{uchida2017embedding,fan2019rethinking,tang2020deep}. For example, double quotation mark `$``$' introduced in Sec.~4.3 can be regarded as a watermark. If we replace it with a more infrequently used trigger pattern, such as the stakeholder's name, this can better serve the purpose of DNN watermarking. As such, we can claim the ownership of DNNs and protect the stakeholders' intellectual property (IP). 

\section{Generalizable New Knowledge Beyond the Narrow World of NLU tasks?}
Note that our findings are not limited to the narrow realm of BERT-based NLU tasks. The findings can be extended to explain and mitigate the shortcut learning problem of other language or language-vision tasks, such as question answering~\cite{mudrakarta2018did}, reading comprehension~\cite{si2019does}, VQA~\cite{agrawal2018don}, etc. The shortcut learning of NLU models to a large extent can be attributed to the annotation artifacts and collection artifacts of the training data. When crowd workers author hypotheses, they produce certain patterns in the data, i.e., annotation artifacts. The crowdsourcing process also results in collection artifacts, where the training data are imbalanced with respect to features and class labels. Both artifacts are not limited to the NLU tasks. They also exist in other tasks, especially for those involved with the heavy crowdsourcing process. These artifacts result in a skewed and long-tailed training set. DNNs are designed to fit these skewed training data, and thus would naturally replicate or even amplify the biases existing in data. Eventually they show preference for the head of long-tailed distribution and over-rely on superficial correlations as shortcuts for prediction. Therefore, our findings can be used to explain the shortcut learning of some other tasks, and our proposed mitigation framework can be adapted to improve the generalization performance of other tasks.


\end{document}